\documentclass[conference]{IEEEtran}
\IEEEoverridecommandlockouts

\usepackage{cite}
\usepackage{amsmath,amssymb,amsfonts}
\usepackage{graphicx}
\usepackage{textcomp}
\usepackage{xcolor}
\usepackage{subfigure}
\usepackage{multirow}
\usepackage{booktabs}
\usepackage{hyperref}
\usepackage{algorithmicx}
\usepackage{algpseudocode}
\usepackage[ruled,linesnumbered]{algorithm2e}

\def\BibTeX{{\rm B\kern-.05em{\sc i\kern-.025em b}\kern-.08em
    T\kern-.1667em\lower.7ex\hbox{E}\kern-.125emX}}
\begin{document}

\makeatletter
\newcommand{\rmnum}[1]{\romannumeral #1}
\newcommand{\Rmnum}[1]{\expandafter\@slowromancap\romannumeral #1@}
\makeatother

\title{USV-AUV Collaboration Framework for Underwater Tasks under Extreme Sea Conditions
}

\author{
\IEEEauthorblockN{
Jingzehua Xu\IEEEauthorrefmark{1}$^{,+}$,
Guanwen Xie\IEEEauthorrefmark{1}$^{,+}$,
Xinqi Wang\IEEEauthorrefmark{2},
Yimian Ding\IEEEauthorrefmark{1},
Shuai Zhang\IEEEauthorrefmark{3}
}
\IEEEauthorblockA{\IEEEauthorrefmark{1}Tsinghua Shenzhen International Graduate School, Tsinghua University, China}

\IEEEauthorblockA{\IEEEauthorrefmark{2}
College of Information Science \& Electronic Engineering, Zhejiang University, China}

\IEEEauthorblockA{\IEEEauthorrefmark{3}Department of Data Science,
New Jersey Institute of Technology, USA}
Email: sz457@njit.edu
\thanks{$^{+}$ These authors contributed equally to this work. 

$^{1}$$\;$Codes are available at https://github.com/360ZMEM/USV-AUV-colab} 
}

\maketitle

\begin{abstract}
    Autonomous underwater vehicles (AUVs) are valuable for ocean exploration due to their flexibility and ability to carry communication and detection units. Nevertheless, AUVs alone often face challenges in harsh and extreme sea conditions. This study introduces a unmanned surface vehicle (USV)–AUV collaboration framework, which includes high-precision multi-AUV positioning using USV path planning via Fisher information matrix optimization and reinforcement learning for multi-AUV cooperative tasks. Applied to a multi-AUV underwater data collection task scenario, extensive simulations validate the framework's feasibility and superior performance, highlighting exceptional coordination and robustness under extreme sea conditions. To accelerate relevant research in this field, we have made simulation code (demo version) available as open-source$^{1}$.
    \end{abstract}
    \section{Introduction}
    \label{sec:intro}
    Autonomous underwater vehicles (AUVs), especially those multi-AUV systems, have attracted considerable interest for their versatility and mobility in applications such as environmental monitoring, seabed exploration, and biological research \cite{1,2}. The effectiveness of these vehicles in underwater tasks is largely dependent on precise and reliable positioning and highly efficient control strategies \cite{3}, particularly in challenging and extreme sea conditions \cite{4}. Accurate positioning is crucial as it boosts operational efficiency and safety by preventing potential damage or accidents. Moreover, AUVs equipped with adaptive control policies \cite{5} and autonomous decision-making capabilities are better suited to handle dynamic environments \cite{6}, providing enhanced resistance to interference and increased practical value \cite{7}.
    
    Current positioning techniques for AUVs mainly consist of global positioning systems, inertial navigation systems, and ultrasonic positioning. However, these methods are vulnerable to problems like water refraction and scattering, signal attenuation, transmission delays \cite{3,8,18}, and error accumulation \cite{5}. In addition, traditional control strategies for multi-AUV systems depend heavily on mathematical models or model-based algorithms \cite{10}. Model-based approaches face difficulties when controller parameters shift in dynamic environments \cite{9}, and both algorithms often struggle to predict the future maneuvers and behaviors of non-cooperative targets, thereby limiting their scalability and adaptability \cite{10,19}.
    
    To address the aforementioned challenges, numerous researchers have concentrated on the development of USV-AUV co-localization and intelligent multi-AUV collaboration. Bahr \textit{et al}. introduced a distributed algorithm that leverages multi-AUV to dynamically determine the locally optimal position of the beacon vehicle, utilizing data from the survey vehicle's broadcast communication \cite{11}. Vasilijevi \textit{et al}. constructed the Internet of underwater things using unmanned surface vehicles (USVs) to enhance the underwater positioning efficiency \cite{12}. Jiang \textit{et al}. applied a multi-agent proximal policy optimization reinforcement learning (RL) algorithm to direct efficient and energy-saving data collection for AUVs in unknown environments based on an objective uncertainty map \cite{14}. Wang \textit{et al}. proposed a collaborative data collection strategy for multi-AUV employing local-global deep Q-learning and data value, to facilitate hybrid data collection meeting various temporal demands \cite{15}. Nevertheless, these methods encounter limitations in complex environments: as number of AUVs and sensors increases, the underwater acoustic channel becomes more intricate, computational complexity rises \cite{13}, and there are stringent requirements for battery and operational costs of 
the equipment. Furthermore, traditional AUV swarm control techniques such as heuristic algorithms \cite{16}, neural networks \cite{7}, and game theory \cite{17}, although effective for specific tasks, heavily rely on extensive prior information. In the absence of such information, performance of these methods significantly declines, particularly under extreme sea conditions.
    

    Based on the above analysis, in this study we propose a USV-AUV collaboration framework for underwater tasks to improve the performance of the AUV completing tasks under extreme sea conditions. To make a conclusion, the contributions of this paper include the following:
    \begin{itemize}
    \item  We realize accurate positioning of AUVs via
    USV path planning through minimizing the determinant of the Fisher information matrix (FIM) of the system. Based on this, through integrating environment-aware ability into state space, and USV-AUV collaboration into reward function in the Markov decision process (MDP), we further use RL to empower multi-AUV with intelligence and the adaptability to extreme sea conditions. 
    
    \item We innovatively leverage two-dimensional tidal wave equations and ocean turbulence model to simulate the extreme sea conditions, which has impact on the positioning accuracy and working efficiency of multi-AUV.
    
    \item 
    Through comprehensive experiments in the underwater data collection task, our framework showcases superior feasibility and excellent performance in balancing multi-objective optimization under extreme sea conditions.
    \end{itemize}
    
    \section{Methodology}
    \label{sec:format}
    In this section, we introduce the proposed USV-AUV collaboration framework, which comprises two main components: high-precision location of multi-AUV using USV path planning via FIM optimization, and RL enabled multi-AUV cooperative work. Besides, we also present simulation of extreme sea conditions using two-dimensional shallow water equation and ocean turbulence model.
    
    \subsection{USV Path Planning Based on Fisher Information Matrix Optimization}
    Our framework realizes accurate positioning of AUVs via USV path planning through minimizing the determinant of the system's FIM. Central to this consensus is that FIM's determinant is negatively correlated to the system's uncertainty.
    
    Assume the coordinate of the USV is denoted as $(x, y, \eta)$, and the coordinates of $k$-th AUV is $(x_k
    , y_k, z_k)$.  As illustrated in Fig. 1, the USV, which is positioned on the sea surface, employs an ultra-short baseline (USBL) system to determine the location of the underwater AUV. The arrays are uniformly spaced on the USV such that the distances $OX\!a \!=\! OX\!b \!=\! OY\!b \!=\! OY\!a \!=\! d/2$, where $d$ represents the array spacing. Consequently, the probability density function of the measured data and the measurement equation can be expressed as follows:
    \vspace{-4mm}
    
    \begin{figure}[!t]
        \centering
        \includegraphics[width=0.948\linewidth]{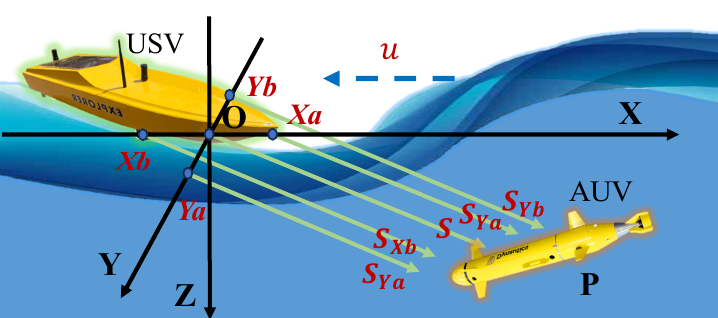}
        \vspace{-1.5mm}
        \caption{Illustration of the AUV positioning via USV. The USV on the sea surface uses USBL to locate the underwater AUV. }
        \label{fig_1}
        \end{figure}
    
    \begin{subequations}
    \begin{align}
    p\!\left(\!\mathbf{\mathit{\boldsymbol{Z}}}\!,\!\mathbf{\mathit{\boldsymbol{X}}}\!\right) \!\!=\!\! \prod_{k = 1}^{m}\!\! &\frac{\mathrm{exp} \!\left\{\!-\frac{1}{2}[ \mathbf{\mathit{\boldsymbol{Z}}}_{k} \!\!-\!\! \mathbf{\mathit{\boldsymbol{h}}}_{k}\left(\!\mathbf{\mathit{\boldsymbol{X}}}\!\right)\left]\right.^{T}\right.\left.\!\!\!\!\mathbf{\mathit{\boldsymbol{R}}}^{- 1}\!\!\left[\mathbf{\mathit{\boldsymbol{Z}}}_{k} \!\!-\!\! \mathbf{\mathit{\boldsymbol{h}}}_{k}\!\left(\!\mathbf{\mathit{\boldsymbol{X}}}\!\right)\right] \right\} }{\sqrt{2 \pi \rm det\left(\mathbf{\mathit{\boldsymbol{R}}}\right)}}\!,\\
    &\!\!\qquad\quad\mathbf{\mathit{\boldsymbol{Z}}}_{k} = \mathbf{\mathit{\boldsymbol{h}}}_{k}\left(\mathbf{\mathit{\boldsymbol{X}}}\right) + \mathbf{\mathit{\boldsymbol{u}}}_{k},\\
    &\!\!\!\! S_{k} = \sqrt{\left(x_{k} - x\right)^{2} + \left(y_{k} - y\right)^{2} + z_{k}^{2}},
    \end{align}
    \end{subequations}
    where the target state vector is denoted by $\mathit{\boldsymbol{X}}\!\!=\!\!\left[x,y\right]^{T}$, while $\mathbf{\mathit{\boldsymbol{h}}}_{k}\!\left(\mathbf{\mathit{\boldsymbol{X}}}\right)\!\!=\!\!\left[\right. \Delta \varphi_{x , k} , \Delta \varphi_{y , k} \left]\right.^{T}$ stands for the phase difference vector between receiving units, including two elements $\Delta \varphi_{x , k} \!\!=\!\! \frac{2 \pi f d}{cS_k}(x_k\!-\!x)$ and $\Delta \varphi_{y , k} \!\!=\!\! \frac{2 \pi f d}{cS_k}(y_k\!-\!y)$, while $c$ represents the speed of sound and $f$ indicates the signal frequency. Additionally, $\mathbf{\mathit{\boldsymbol{u}}}_{k}$ is characterized as zero-mean Gaussian white noise, and the measurement noise covariance matrix is denoted by $\mathbf{\mathit{\boldsymbol{R}}}\!\!=\!\!\sigma^{2}\mathit{\boldsymbol{I}}$.

    \begin{figure}[!t]
        \centering
        \includegraphics[width=0.948\linewidth]{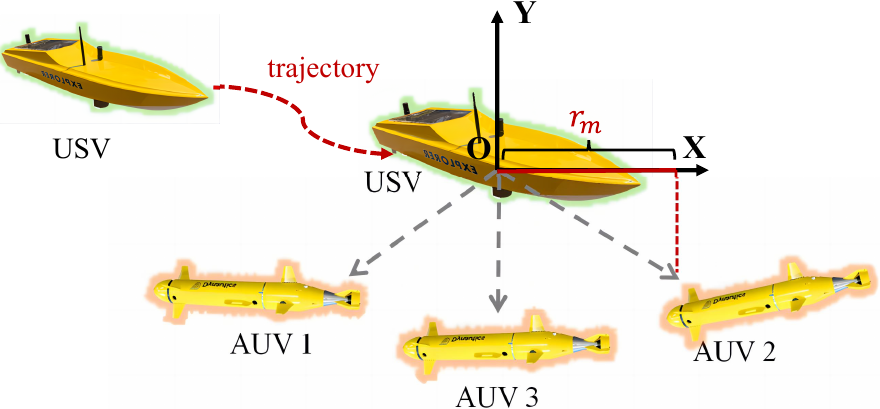}
        \vspace{-1.5mm}
        \caption{The diagram of USV path planning. We can realize accurate positioning of AUVs via USV path planning through minimizing the determinant of the system’s FIM.}
        \label{fig_2}
        \end{figure}
    
    Then FIM of the system is subsequently obtained by determining the second derivative of the log-likelihood function

    \vspace{-4mm}
    \begin{equation}
    \begin{aligned}
    \mathbf{\mathit{\boldsymbol{J}}}_{m} &= \left[\begin{array}{ccc}- E \left[\frac{\partial^{2} {\rm l n} p \left(\mathbf{\mathit{\boldsymbol{Z}}} , \mathbf{\mathit{\boldsymbol{X}}}\right)}{\partial x^{2}}\right]  & - E \left[\frac{\partial^{2} {\rm l n} p \left(\mathbf{\mathit{\boldsymbol{Z}}} , \mathbf{\mathit{\boldsymbol{X}}}\right)}{\partial x\partial y}\right] \\
    - E \left[\frac{\partial^{2} {\rm l n} p \left(\mathbf{\mathit{\boldsymbol{\boldsymbol{Z}}}} , \mathbf{\mathit{\boldsymbol{X}}}\right)}{\partial y\partial x}\right] & - E \left[\frac{\partial^{2} {\rm l n} p \left(\mathbf{\mathit{\boldsymbol{Z}}} , \mathbf{\mathit{\boldsymbol{X}}}\right)}{\partial y^{2}}\right]\end{array}\right].
    \end{aligned}
    \end{equation}
    
    Assume there are totally $m$ AUVs, the determinant of the FIM can be simplified to the final expression after derivation, which can be expressed as

    \vspace{-4mm}
    \begin{subequations}
    \begin{align} {\rm det}\!\left(\!\mathbf{\mathit{\boldsymbol{J}}}_{m}\!\right) \!= \!&\left(\!\right.\! \frac{4 \pi^{2} f^{2} d^{2}\!}{\sigma^{2} c^{2}} \left.\!\right)^{2} \!\!\left[3 m \frac{{\rm sin}^{2} \gamma_{0}}{S_{0}^{4}} \!+\! \frac{ \left({\rm sin}^{4} \gamma_{0} \!+\! 1 \right)^{2}}{S_{0}^{4}}\chi\right]\!,\\
    &\qquad\! r_m = \rm argmax \left\{d e t \left(\mathbf{\mathit{\boldsymbol{J}}}_{m}\right)\right\},
    \end{align}
    \end{subequations}
    where ${\rm sin}\gamma_0\!\!=\!\!\frac{z_k}{S_0}$, while $\chi\!\!=\!\!\sum_{1\leq i \textless j\leq m}^m {\rm \!sin}^{2}\alpha_{ij}$, with $\alpha_{ij}\!\!=\!\!\varphi_i\!-\!\varphi_j$ representing the angle between the projections of two AUVs and USV. By maximizing the determinant ${\rm det}\!\left(\!\mathbf{\mathit{\boldsymbol{J}}}_{m}\!\right)$, we can determine the optimal horizontal distance $r_m$ between the USV and multiple AUVs, as illustrated in Fig. 2.
    
    \subsection{Reinforcement Learning Enabled Multi-AUV Collaboration}
    Our framework leverages RL to train multi-AUV for collaborative operations. Rather than solely relying on standard MDP based RL algorithms, we introduce modifications to design of the state space and reward functions in standard MDP. Specifically, we incorporate the ocean current velocity perceived by AUV $k$ into its state space, represented as
    \vspace{-2mm}
    
    \begin{equation}
    s_k=\left\|\boldsymbol{V}_c\left ( \boldsymbol{P}_k(t) \right )\right\|.
    \end{equation}
    
    Furthermore, we integrate original reward functions with the distance differential between each AUV and the USV, which can be denoted as
    \vspace{-2mm}
    \begin{equation}
    r_{k}\!(t)=\left(l_{\max }^{k \leftrightarrow U} / l^{k \leftrightarrow U}(t)\right),
    \end{equation}
    where $l_{\max }^{k \leftrightarrow U}$ and $l^{k \leftrightarrow U}$ stand for the maximum distance, and indicates the current distance between AUV $k$ and USV. Through extensive epochs of RL training, the collaborative behavior and decision-making capabilities of multi-AUV, enhanced with environment-awareness, will progressively converge to an expert level.

    \begin{algorithm}[!t]
    \label{alg:1}
    \caption{USV-AUV Collaboration Framework}
    Initialize USV-AUV's position, the replay buffer $\mathcal{D}$, critic and actor network parameters of each AUV.
    
    \For{each epoch $k$}{
    Reset the training environment and parameters.
    
    \For{each time step $i$}{
    USV path planning via FIM optimization.
    
    Measure each AUV's position via Eq. (1).
    
    \For{each AUV $k$}{
    Obtain current state information $s_i$.
    
    Sample and execute action $a_i$ while receiving reward $r_i$. 
    
    Transfer to the next state $s_{i+1}$.
    
    \While{In updating period}{
    Extract $N$ tuples of data ${(\boldsymbol{s}_n,\!a_n\!,r_n,\boldsymbol{s}_{n+1})}_{n\!=\! 1,\!\cdots\!,N}$ \!from $\mathcal{D}$.
    
    Update the Critic Nework.
    
    Update the Actor Network.
    
    }
    
    Store $(\boldsymbol{s}_i,a_i,r_i,\boldsymbol{s}_{i+1})$ in $\mathcal{D}$.
    
    }
    }}
    \end{algorithm}

    Combining Sections \Rmnum{2}-A and \Rmnum{2}-B, we finally propose the USV-AUV collaboration framework, whose algorithm pseudo-code has been listed in Algorithm 1.
    
    \subsection{Simulation of Extreme Sea Conditions}
    Given that the USV operates on the surface and the AUVs navigate underwater, the USV-AUV collaboration framework is highly susceptible to disturbances from severe waves and ocean turbulence.
    
    Based on this intuition, our study employs two-dimensional shallow water equations to simulate the sea surface with wave dynamics. If we denote the wave velocity as $\boldsymbol{V}\!_w\!=\![u,v]$, and the gravitational acceleration as $g$, the water level $\eta$ at coordinate point $(x^{'},y^{'}\!)$ can be calculated by
    \begin{subequations}
    \begin{gather}
    \frac{\partial u}{\partial t} + g \frac{\partial \eta}{\partial x^{'}} = 0,\\
    \frac{\partial v}{\partial t} + g \frac{\partial \eta}{\partial y^{'}} = 0,\\
    \frac{\partial \eta}{\partial t} + \frac{\partial \left(u \cdot h\right)}{\partial x^{'}} + \frac{\partial \left(v \cdot h\right)}{\partial y^{'}} = 0,  
    \end{gather}
    \end{subequations}
    where $h$ represents the water depth, and we can further derive the expression
    
\begin{equation}
    \eta =R_{L} \frac{c o s k x^{'}}{c o s k L} e^{- i \omega t},
\end{equation}
where $R_L$ denotes the variable associated with the offshore length $L$, while $k \!\!=\!\! \frac{2 \pi}{\lambda}$, with wave length $\lambda \!\!=\!\!\frac{2 \pi}{\omega} \sqrt{g h}$. Consequently, we observe that $\cos kL$ becomes zero when $L\!\!=\!\!\frac{\lambda}{4}$, resulting in a significant rise in the water level.

    \begin{figure*}[!t]
        \centering
        \subfigure[Sum data rate]{
        \includegraphics[width=0.317\linewidth]{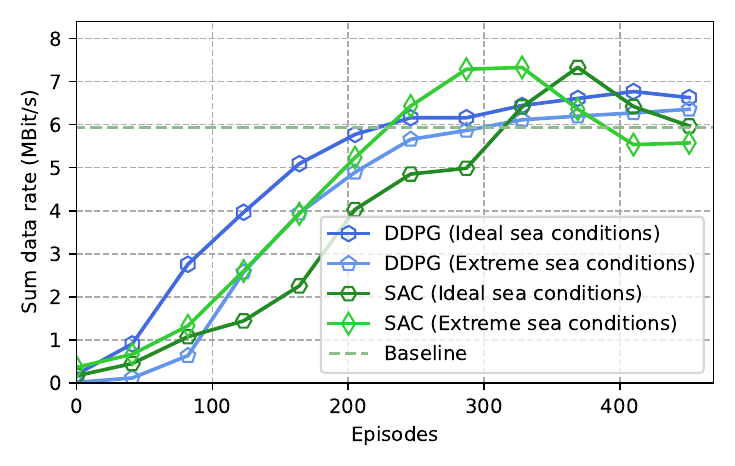}}
        \subfigure[Energy consumption]{
        \includegraphics[width=0.317\linewidth]{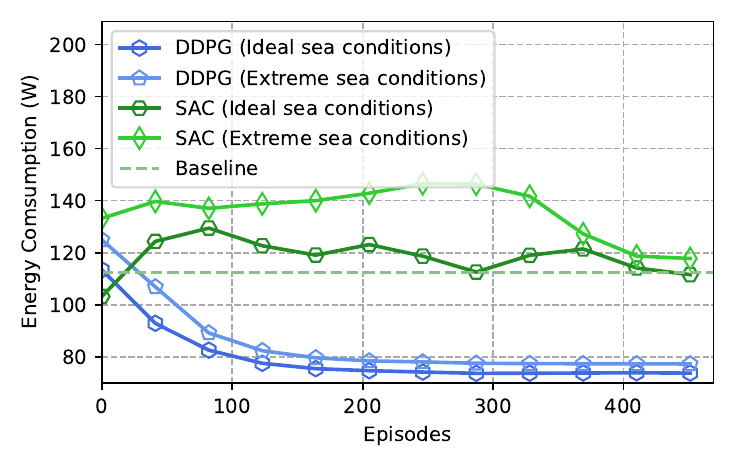}}
        \subfigure[Average reward per timestep]{
        \includegraphics[width=0.317\linewidth]{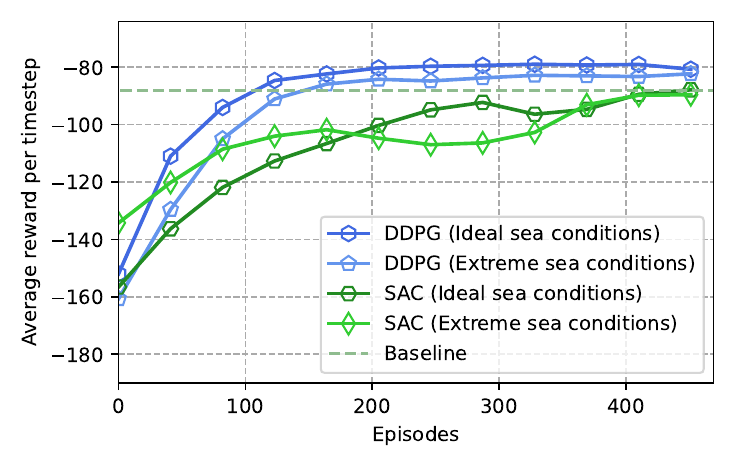}}
        \vspace{-2mm}
        \caption{The curves of sum data rate, energy consumption and average reward per timestep, using DDPG and SAC for RL training in ideal and extreme sea conditions, respectively. (a) Sum data rate. (b) Energy consumption. (c) Average reward per timestep.}
        \vspace{-4mm}
        \label{fig_3}
        \end{figure*}
    
    Additionally, this study employs the superposition of multiple viscous vortex functions, derived from the simplified Navier-Stokes equations, to simulate ocean turbulence. The functions are presented as follows:
    \begin{subequations}
    \begin{gather}
    V_x(\!\boldsymbol{P}_k(t))\!\!=\!\!-\!\Gamma \!\cdot \frac{y^{'}-y_0}{2 \pi\left\|\boldsymbol{P}_k(t)\!\!-\!\!\!\boldsymbol{P}_0\right\|_2^2} \!\!\cdot\!\!\left(\! 1\!-\! e^{\!-\!\frac{\left\|\boldsymbol{P}_k(t)-\!\boldsymbol{P}_0\right\|_2^2}{\delta^2}}\!\right)\!\!,\label{eq:12a}\\
    V_y(\!\boldsymbol{P}_k(t))\!\!=\!\!-\!\Gamma \!\cdot \frac{x^{'}\!-\! x_0}{2 \pi\left\|\boldsymbol{P}_k(t)\!\!-\!\!\!\boldsymbol{P}_0\right\|_2^2} \!\!\cdot\!\!\left(\! 1\!-\! e^{-\!\frac{\left\|\boldsymbol{P}_k(t)-\!\boldsymbol{P}_0\right\|_2^2}{\delta^2}}\!\right)\!\!,\label{eq:12b}\\
    \varpi(\boldsymbol{P}_k(t))=\frac{\Gamma}{\pi \delta^2} \cdot e^{-\frac{\left\|\boldsymbol{P}_k(t)-\boldsymbol{P}_0\right\|_2^2}{\delta^2}},
    \end{gather}
    \end{subequations}
    where $\boldsymbol{P}_k(t)$ and $\boldsymbol{P}_0$ represent the current location of the AUV $k$ and the coordinate vector of Lamb vortex center, respectively. $V_x(\boldsymbol{P}_k(t))$ and $V_y(\boldsymbol{P}_k(t))$ are the velocities of the ocean turbulence on the $X$ and $Y$ axes perceived by AUV $k$ at position $\boldsymbol{P}_k(t)$ at time $t$, respectively. Meanwhile, $\varpi$, $\delta$, and $\Gamma$ stand for the vorticity, radius, and intensity of the vortex, respectively.
    
    Finally, we can employ the finite difference method to simulate sea waves and the ocean turbulence model, which together constitute the extreme sea conditions in this study.
    
    \section{Experiments}
    \label{sec:exp}
    In this section, we verify the effectiveness of the proposed USV-AUV collaboration framework under extreme sea conditions through comprehensive simulation experiments. Furthermore, we present the experimental results with further analysis and discussion.
    
    \subsection{Task Description and Settings} 
    \label{subsec:fc}
    Since open-source underwater tasks are scarce, we have chosen the multi-AUV data collection task as a representative example to evaluate our framework. This task involves utilizing multi-AUV to gather data from underwater sensor nodes within the Internet of underwater things (IoUT), with multiple objectives such as maximizing the total data rate, avoiding collisions, and minimizing energy consumption, etc. The parameters used in this paper are detailed in Table 1. For additional detail and parameters related to the task, please refer to the previous work \cite{4}.

    \subsection{Experiment Results and Analysis} 
    To evaluate the feasibility of our framework, we first employed two mainstream RL algorithms, DDPG and SAC, to train two AUVs positioned by the USV on the sea surface to collaboratively complete the underwater data collection task under both ideal and extreme sea conditions, respectively. As illustrated in Fig. 3, the training curves progressively converge to expert-level performance, indicating that the AUVs have successfully acquired the expert policy through RL training. Additionally, we performed environmental generalization experiments to compare performance in both ideal and extreme sea conditions. The corresponding results (here ISC, ESC denote ideal and extreme sea conditions, while SDR, EC, ARPS indicate sum data rate, energy consumption, and average reward per timestep, respectively), presented in Table 2, show that despite the presence of ocean waves and turbulence, the performance remains comparable in both scenarios. This demonstrates the framework's high robustness under extreme sea conditions.
    
    \begin{table}[!t]
        \centering
        \caption{\!Parameters of the environment and proposed\\framework in this study.}
        \label{table:1}
        \vspace{-2mm}
        \begin{tabular}{lc} 
        \toprule
        Parameters & Values \\
        \midrule
        Experimental area    & 200m $\times$ 200m   \\
        USV sound level  & 135dB  \\ 
        USV transmit frequency & 15kHz(AUV\! 1)/18kHz(AUV\! 2) \\
        Hydrophone array $d$ & 0.033m \\
        Water Depth & 120m \\ 
        Time step $\Delta t$  & 10s \\
        Space step $\Delta x$ & 4m \\ 
        Angular frequency $\omega$ & 2$\pi$/43200 rad/s \\ 
        Initial amplitude $\eta_0$ & 5m \\ 
        \bottomrule
        \vspace{-4mm}
        \end{tabular}
        \end{table}

  \begin{table}[!t]
            \centering
            \caption{Comparison of different RL algorithms under ideal and extreme sea conditions, respectively.}
            \label{table:2}
            \vspace{-2mm}
            \begin{tabular}{lcccc}
            \toprule
              & SDR & EC & ARPS \\
            \midrule
            DDPG (ISC)         & 6.63±0.97 & 73.72±1.11 & -80.70±6.94 \\
            DDPG (ESC)      & 6.35±1.57  & 77.23±0.89  & -82.28±7.41  \\
            SAC (ISC)  & 5.97±2.66   & 111.57±13.23  & -87.86±13.10  \\
            SAC (ESC)    & 5.58±2.35  & 117.76±17.30 & -89.58±15.22 \\
            Baseline \cite{4}  & 5.93±3.51  & 112.30±10.56 & -87.95±12.08 \\
            \bottomrule
            \vspace{-6mm}
            \end{tabular}
        \end{table}

    Moreover, we utilized the expert policy trained with the DDPG algorithm to guide the multi-AUV system in an underwater data collection task under extreme sea conditions. The trajectories of the AUVs and the USV during a single RL training episode are depicted in Fig. 4(a). To further evaluate the advantages of the USV-AUV collaboration framework, we also presented the positioning error of the multi-AUV system. We examined three different scenarios: employing USV path planning based on FIM optimization, fixing the USV at coordinates (0,0), and fixing it at (100,100). As shown in Fig. 4(b), the first approach achieves the lowest positioning error, illustrating the superior performance of USBL positioning via USV path planning using FIM optimization, even under extreme sea conditions.

 \begin{figure}[!t]
        \centering
        \subfigure[Trajectories of AUVs and USV]{
        \includegraphics[width=0.48\linewidth]{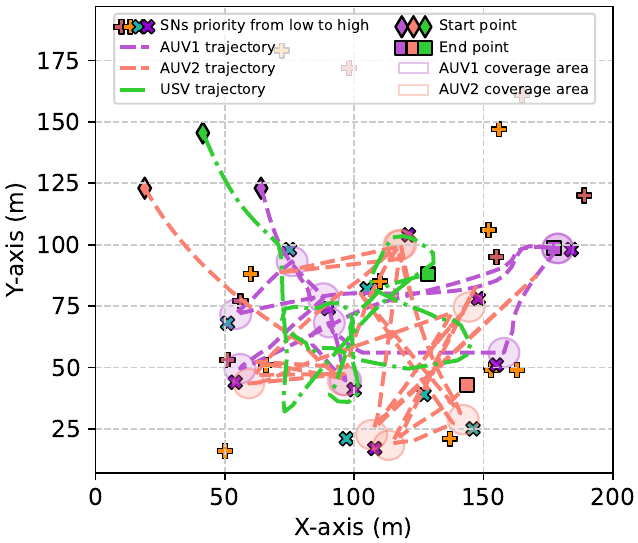}}
        \subfigure[Positioning error of the AUV]{
        \includegraphics[width=0.48\linewidth]{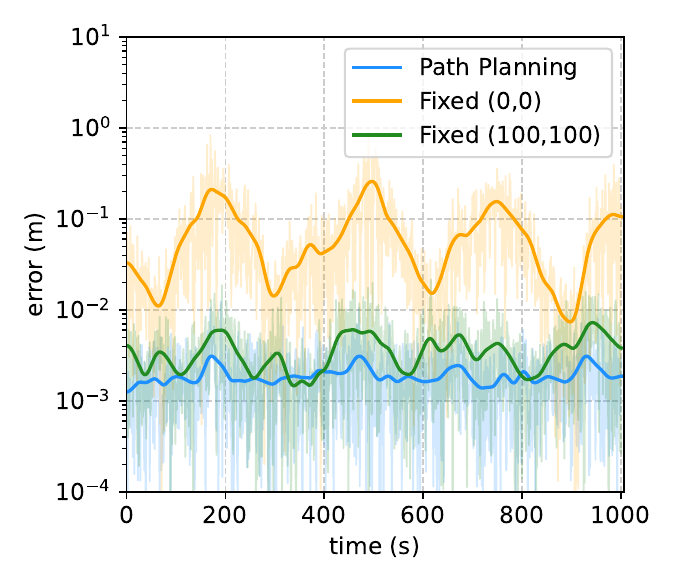}}

        \vspace{-1.5mm}
        \caption{Trajectories of AUVs and USV, and positioning error of the AUV with USV fixed at (0,0) and (100, 100), and path planning using FIM optimization, respectively. (a) Trajectories of AUVs and USV. (b) Positioning error of the AUV.}
        \label{fig_6}
        \vspace{-3mm}
        \end{figure}
        \label{subsec:mr}


    \section{Conclusion} 
    In this paper, we propose a USV–AUV collaboration framework. The two parts of our framework, including high-precision
    multi-AUV location using USV path planning by FIM optimization and RL training
    for multi-AUV cooperative tasks, collaboratively enhance the performance of multi-AUV underwater tasks in extreme sea conditions. Experiment results from underwater data collection task verify the superior feasibility and performance of the proposed framework, which effectively embodies high coordination between USV and AUV while showcasing excellent robustness in extreme sea conditions. To accelerate relevant research in this field, we have made simulation codes (demo version) available as open-source at https://github.com/360ZMEM/USV-AUV-colab .
    \label{sec:con}

    \section{Acknoledgments} 
    Part of this work was done when Jingzehua Xu was studying at Tsinghua University, and in the MicroMasters Program in Statistics and Data Science at Massachusetts Institute of Technology (MIT), respectively. We would like to express our heartfelt gratitude to Yong Ren at Tsinghua University for his generous praise. Our sincere appreciation also goes to Prof. Shuai Zhang at the New Jersey Institute of Technology (NJIT) for his unwavering support. Furthermore, we genuinely appreciate the encouragement from Prof. Miao Liu and Prof. Songtao Lu at IBM Research, as we have gained numerous valuable insights from their previous work, and their positive feedback has greatly inspired us to move forward in multi-agent systems.

    \bibliographystyle{IEEEtran}


\end{document}